\documentclass[twocolumn]{article}  
 
\usepackage{arxiv, amsmath}
 
\usepackage[normalem]{ulem}
\usepackage{amsmath}
\usepackage[utf8]{inputenc} 
\usepackage[T1]{fontenc}
\usepackage{cite}
\usepackage[colorlinks=false, linkcolor=black, citecolor=black, urlcolor=black]{hyperref}
\usepackage{hyperref}       
\usepackage{url}            
\usepackage{booktabs}       
\usepackage{amsfonts}       
\usepackage{nicefrac}
\usepackage{authblk}
\usepackage{microtype}
\usepackage{lipsum}
\usepackage{graphicx}
\graphicspath{{./images/}}
\usepackage[title]{appendix}
\usepackage[section]{placeins}
\usepackage{import}
\usepackage{lipsum,mathtools,cuted}
\usepackage{caption}
\DeclareCaptionFont{xxviii}{\fontsize{8}{0}\selectfont}
\usepackage{abstract}
 \usepackage{appendix} 
\begin{document}
 
\title{The advantage of fine-grained training}
\author[1]{Davide Pirovano}
\author[1]{Federico Milanesio}
\author[1]{Michele Caselle}
\author[2]{Piero Fariselli}
\author[  ]{Matteo Osella (corresponding author)\textsuperscript{1,}\thanks{Corresponding author email: \texttt{matteo.osella@unito.it}}}
 
\affil[1]{Department of Physics and INFN, University of Turin, Via Giuria 1, 10125 Turin, Italy}
\affil[2]{Department of Medical Sciences, University of Turin, Via Santena 19, 10123 Turin, Italy}
 
\date{November 25, 2025}
 
\twocolumn[
    \maketitle
    \begin{abstract}
    
In classification problems, models {are trained to} predict a class label based on the input data features. However, class labels are organized hierarchically in many datasets. While a classification task is often defined at a specific level of this hierarchy, training can utilize a finer granularity of labels. Empirical evidence suggests that such fine-grained training can enhance performance. In this work, we investigate the generality of this observation and explore its underlying causes using both real and synthetic datasets. We show that training on fine-grained labels does not universally improve classification accuracy. Instead, the effectiveness of this strategy depends on the geometric structure of the data and its relations with the label hierarchy. {Specifically, we show that the advantage of fine-grained training crucially depends on the degree of alignment between the decision boundaries required for the fine- and coarse-grained tasks, a property that we term boundary redundancy.}
Additionally, factors such as dataset size and model capacity significantly influence whether fine-grained labels provide a performance benefit. {Indeed, we identify a transition,  whose location is largely controlled by the degree of overparameterization, separating regimes where fine-grained training improves performance from those where direct coarse-grained training is preferable.}
    
    \end{abstract}
    \vspace{1cm}
]
\saythanks
\section{Introduction}
 
\begin{figure*}
    \centering
    \includegraphics[width=0.6\linewidth]{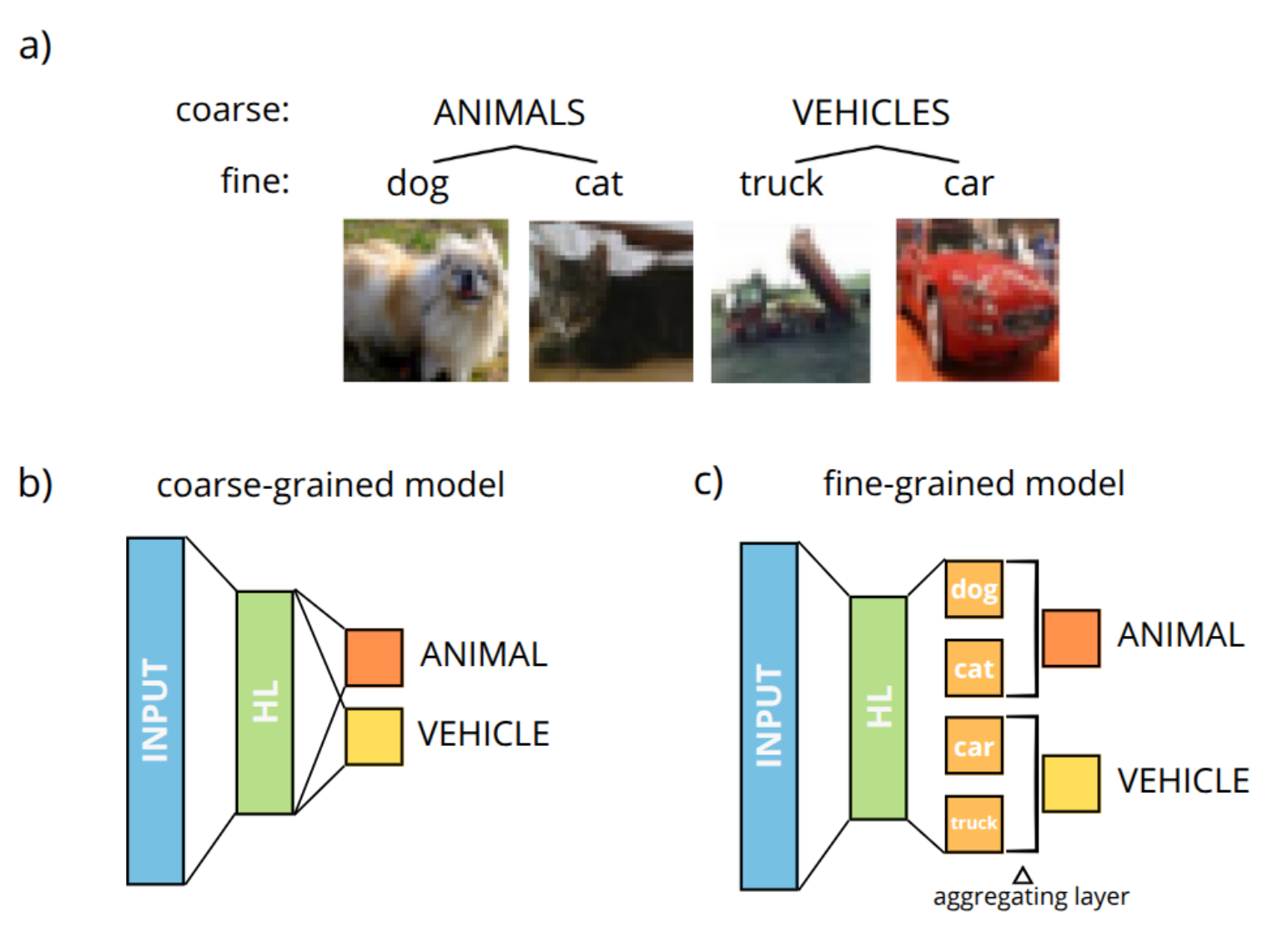}
    \caption{\textbf{Coarse and fine labeling} (a) A possible way to categorize four CIFAR-10 classes (dogs, cats, cars, and trucks) into two broad coarse classes (animals and vehicles). Two models are employed during training: (b) coarse-grained and (c) fine-grained. The coarse-grained model is trained to differentiate between broad categories ({for example}, animal vs. vehicle in the image). Instead, the fine-grained model is trained to classify specific subclasses within those categories. During testing, the fine-grained model is adapted to predict coarse classes by introducing an \textit{aggregating layer} (shown in orange in the figure). This layer sums the output probabilities of the fine-grained model into binary predictions, aligning with the coarse class structure. 
}
 
    \label{fig:dataset_subdivision}
\end{figure*}
 
In supervised machine learning,  labeled data are used to train a model to perform a specific task~\cite{lecun_deep_2015, prince_understanding}.
The definition of a label depends on the task, but data points are often associated with multiple possible labels at different levels of granularity. 
For example, an image of a cat could be labeled as \textit{animal},  \textit{mammal}, \textit{pet}, and so on, and a classification task could be defined at any of these levels. 
 
Consider a dataset containing images of dogs, cats, trucks, and cars (Fig.~\ref{fig:dataset_subdivision}a): we can be interested in the broad separation between animals and vehicles or the finer distinction between the four classes. This decision is set by the required level of detail needed for our application, but during the training phase, we can use all labels at our disposal. 
This situation is typical of classification problems, as many categorical datasets are inherently organized in hierarchical structures~\cite{taxonomy}, often described by a tree or a directed acyclic graph~\cite{silla2011survey}.
 
For example, widely used databases like ImageNet~\cite{deng2009imagenet} employ a hierarchical labeling scheme derived from semantic relationships defined in natural language processing, specifically in the WordNet database~\cite{miller1995wordnet}. Beyond computer vision, hierarchical structures of categories appear in numerous domains. Notable examples include the classic Linnaean classification of species in biology,  the International Classification of Diseases (ICD) maintained by the World Health Organization (WHO), and the Gene Ontology database~\cite{gene2023gene},  which assigns functional annotations to genes at multiple resolution levels.

{Because of this ubiquitous hierarchical data structure, several machine learning approaches have explicitly exploited it during training or prediction~\cite{silla2011survey}. Here we focus on perhaps the simplest strategy: training  on fine-grained labels even when the task is defined at a coarser level. 
}
Given a classification task,  a straightforward labeling choice is to align the label granularity to the task. For example, if the goal is to learn to separate images of animals and vehicles, we can label the training data using these two broad, coarse classes. 
This seems a natural choice since there is no need to separate the type of vehicle - a car or a ship - in this application.
However, an alternative approach would be to train the model using fine-grained labels, forcing it to differentiate between specific types of animals or vehicles, even if the model's performance will be evaluated only in terms of the coarse classes. The model prediction on coarse classes can be trivially obtained by hierarchically aggregating its predictions on fine labels. Figure~\ref{fig:dataset_subdivision}b presents a sketch of these two alternative training strategies for a neural network model. 
 
Empirical evidence suggests that fine-grained training {improves model performance} in separating coarse-grained classes. For example, this approach played a key role in a potentially transformative application of deep learning for classifying benign versus malignant skin lesions using images taken with cell phones~\cite{dermatologist}. In that case, training a large neural network on highly specific disease annotations was crucial for achieving performance on par with human dermatologists in the clinically relevant binary classification task. More generally, experiments on standard image classification datasets indicate that neural networks can benefit from training (or pre-training followed by fine-tuning) on fine-grained labels~\cite{Chen,ridnik2021imagenet,hong2024}. 
{
Hierarchical information has also been exploited beyond computer vision, for example in natural language processing, where hierarchical representations or decisions have been successfully used for tasks such as text summarization and hierarchical classification~\cite{inan2025making,bhambhoria2023simple}.}
However, the generality of this result and its potential explanations remain largely unclear.

\begin{figure*}[!ht]
    \centering
    \includegraphics[width=0.83\linewidth]{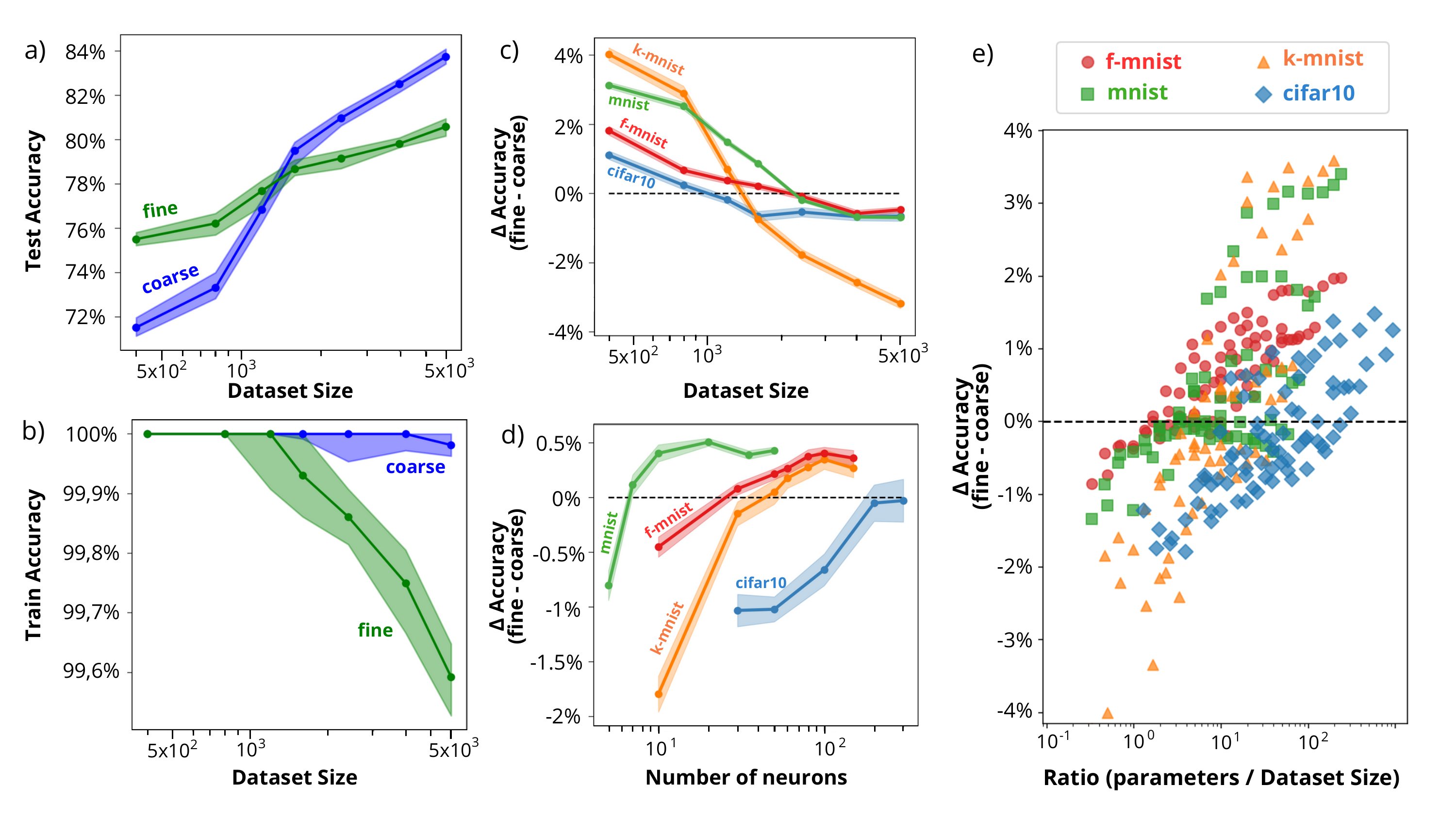}
    \caption{\textbf{Relative performance depends on dataset size and model capacity.}  (a) Test accuracy on K-MNIST as a function of training set size for neural networks trained on fine-grained labels (green) and on coarse-grained labels (blue). (b) Training accuracy for the same models. 
    Both panels use networks with 10 hidden neurons, trained with stochastic gradient descent (SGD). The batch size scales with dataset size (8 for the smallest, 32 for the largest). The learning rate was scheduled to linearly decrease from 0.01 to 0.001. Test set of $10^4$ points, with error bars representing the first and third quartiles of accuracy obtained from 30 independent training runs with different initializations. (c) The difference in test accuracy, $\Delta\text{Accuracy}= \text{Acc}_{\text{fine}}-\text{Acc}_{\text{coarse}}$, across datasets (CIFAR-10, K-MNIST, F-MNIST, MNIST) at different dataset sizes. Networks with 10 hidden neurons for MNIST, F-MNIST, K-MNIST, 60 hidden neurons for CIFAR-10. All were trained with SGD. (d)  $\Delta\text{Accuracy}$ for the same models as a function of the number of neurons at fixed training set sizes (1200 for MNIST; 3200 for the others). Error bars in (c) and (d) are the standard error over 30 runs. (e) $\Delta\text{Accuracy}$ is plotted against the ratio of model parameters to dataset size for the various datasets. Each point represents the fine-grained model's accuracy gain (or loss) for a specific combination of dataset size and model capacity. Coarse groupings are numbers 0 to 3 against numbers 4 to 8 for MNIST, [T-shirt/top, sandal, dress, ankle boot] vs. [pullover, sneaker, shirt, bag] for F-MNIST, [o, ki, su, tsu] vs. [na, ha, ma, ya] for K-MNIST, and vehicles [airplane, automobile, ship, truck] vs. animals [dog, deer, bird, cat] for CIFAR-10.}
    \label{fig:2}
\end{figure*}

In principle, training with fine-grained labels forces the network to learn more specific features that capture the subtleties necessary to distinguish between refined categories. This could prevent the model from \textit{shortcut learning} ~\cite{geirhos2020shortcut}, {namely}, using oversimplistic solutions or even leveraging biases present in the dataset, and thus increase the generalization properties. This is essentially the explanation proposed by previous theoretical work on the benefit of fine-grained pre-training~\cite{hong2024}. 
 
The ability to \textit{compress} data internal representations by focusing exclusively on features useful for the task, while discarding irrelevant information,   
has been linked to the astonishing generalization performance of deep neural networks~\cite{tishby2015deep,achille2018emergence,ciceri2024inversion}. More specifically, 
the network's implicit ability to cluster subclasses in their representations, even without explicit fine-grained labels, seems to correlate with generalization performance across various settings~\cite{carbonnelle2020intraclass}. 
Therefore, it is natural to imagine that supervised fine-grained training can further boost this compression process and improve generalization.
{This intuition is closely related to the guiding principles of multitask learning and auxiliary learning, in which auxiliary tasks can improve generalization by constraining the learned representations and effectively acting as an inductive bias~\cite{caruana1997multitask,crawshaw2020multi}. Training on fine-grained labels can similarly be viewed as introducing an auxiliary  task, {namely}, separating subclasses, in addition to the original coarse classification task.}


However, one can also imagine that in some cases, the data features that fine-training forces the network to learn could be unnecessary for the task defined at the coarse level, and thus could simply represent irrelevant or even distracting information that should be neglected to avoid overfitting.
{
A similar phenomenon has indeed been  observed in auxiliary learning, where the addition of   insufficiently related tasks can degrade the performance of the target task, an effect often referred to as \textit{negative transfer}\cite{vandenhende2021multi}. }

Also in the context of network distillation, {namely}, the process of knowledge transfer from a large model to a smaller one ~\cite{hinton2015distilling}, the transfer often works more efficiently when a large number of labeled classes is considered~\cite{Hinton}, again pointing toward an implicit advantage of a fine-grained class description. The authors indeed explain this effect by the additional information bits that labels describing a finer partition of classes inherently provide.
However, several questions remain open: Under what conditions is this additional information useful? Can it instead act as a confounding factor, and if so, in what cases?
 
This paper tackles these questions and investigates the factors determining the effectiveness of fine-grained training. We demonstrate that such a training strategy does not always provide an advantage; {rather, its utility depends on   the interplay between dataset size and model expressivity, and crucially on the geometric structure of the dataset}. 
{Indeed, the trade-off between the advantage due to the  additional information provided by learning the fine-grained classes  and the increased optimization burden required to separate those classes is  largely determined by data geometry,  and specifically by the degree of alignment between the boundaries required by fine- and coarse-grained tasks. We formalize this geometric intuition using the concept of \textit{boundary redundancy}  and   show that boundary redundancy is indeed a key determinant of  the advantage of fine-grained training.}
{We further assess the robustness of our findings across different datasets and neural network architectures, considering both fully connected and convolutional models.  However, we leave the full exploration of state-of-the-art models and transformer-based architectures for future work. The approach we take  here} is inspired by statistical physics~\cite{zdeborova2020understanding}: we focus on simple models and seek general emerging trends, rather than employing state-of-the-art sophisticated architectures that might hinder the understanding of fundamental principles.

\section{Results} 
\subsection{Fine-grained labels are not always optimal for training}\label{sec:real-world}
 
The first question we investigate is whether training on fine-grained labels consistently improves generalization performance, as suggested by the empirical evidence presented in the introduction ({for example}, refs.~\cite {dermatologist, Chen}). 
To answer this question, we selected several standard datasets, described in Sec.~\ref {methods:datasets} of the Methods, and organized their classes into different hierarchical structures. Specifically,  we define two broad coarse classes, each composed of four fine-grained subclasses, and compare model performance on the coarse binary classification task when trained at the two levels of label granularity. The details of the fine- and coarse-grained architectures are {explained} in the Methods, Sec.~\ref{methods:models}. Figure~\ref{fig:2}a shows this comparison for two neural networks with the architectures shown in Fig.~\ref{fig:dataset_subdivision}b and c, trained on subsamples of the K-MNIST dataset of varying sizes. The model trained on fine-grained labels outperforms the coarse-grained one only for small dataset sizes. In contrast, training the network directly on coarse labels proves to be the better strategy for larger dataset sizes. Therefore, training on fine-grained labels is not universally advantageous.
 
Training on fine-grained labels contributes to performance gains primarily by enhancing the model’s ability to generalize. When the fine-grained model generalizes better, both models still interpolate nearly perfectly on the training set (Fig.~\ref{fig:2}b). However, as the dataset size increases and the fine-grained model begins to be outperformed, we observe a drop in its training accuracy (evaluated on the coarse labels). This suggests that, in certain settings, the more challenging task of distinguishing fine-grained classes can hinder optimization and, as a result, degrade final performance. This observation generalizes to other datasets: coarse-trained networks can match or exceed the training accuracy of their fine-grained counterparts, but tend to overfit more for small dataset sizes, as shown in Supplementary Materials, Fig.~S1. Moreover, in Fig.~S2, we show that this result is robust for different configurations of the coarse F-MNIST datasets.

\begin{figure*}[!ht]
    \centering
    \includegraphics[width=.71\linewidth]{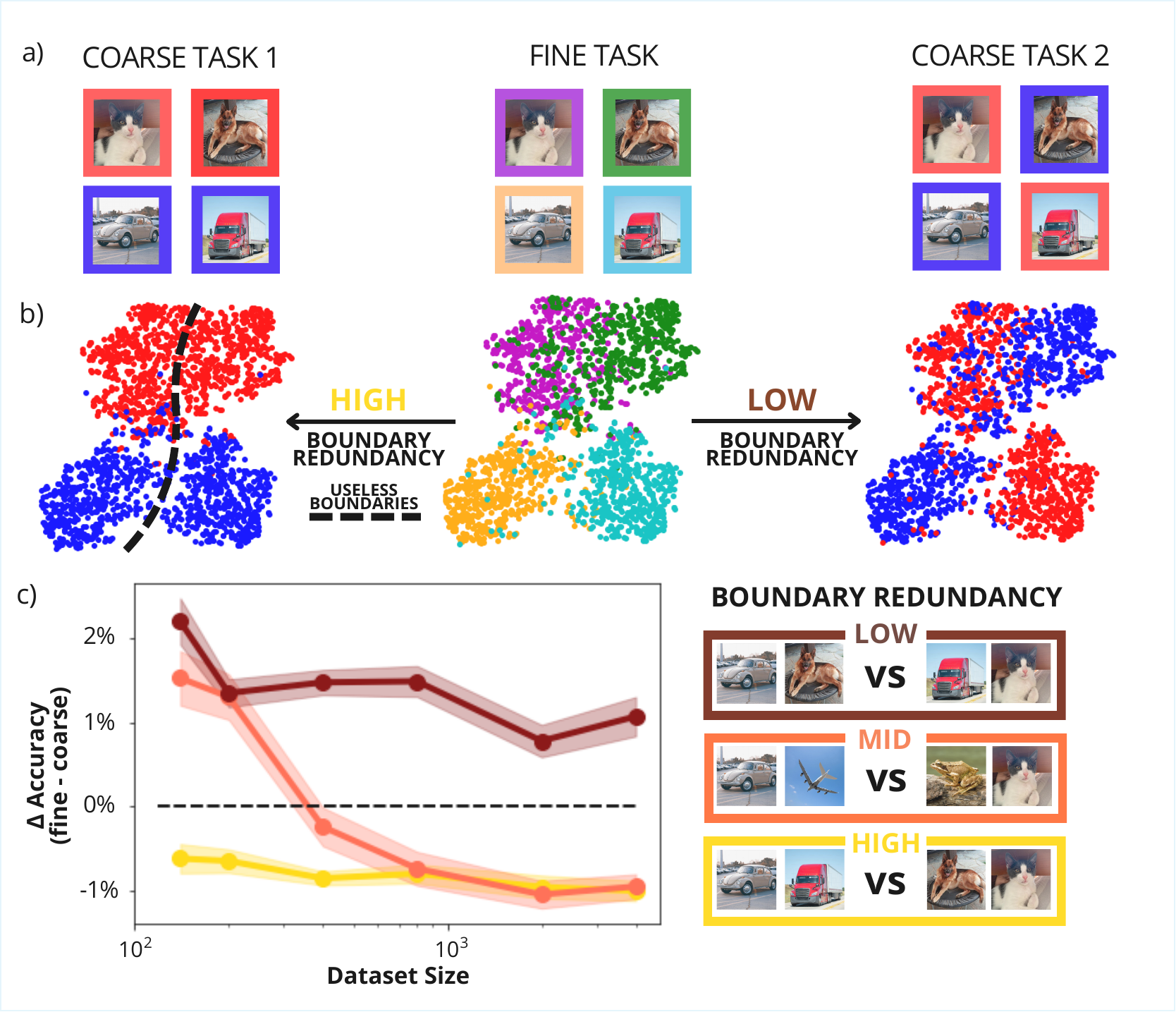}
    \caption{\textbf{The impact of boundary redundancy on generalization.} (a) Four CIFAR-10 classes (\textit{cat}, \textit{dog}, \textit{car}, \textit{truck}) are grouped into two alternative coarse classification tasks. Coarse Task 1: animals vs. vehicles, where subclasses share natural attributes. Coarse Task 2: (\textit{cat}, \textit{truck}) vs. (\textit{dog}, \textit{car}), which mixes dissimilar subclasses. (b) t-SNE embedding of the internal representation learned by a convolutional neural network trained on the fine-grained four-class task (see Methods, Sec.~\ref{methods:tsne}). Points are recolored according to the coarse tasks: red vs. blue for the two coarse classes. In this space, some fine-grained boundaries are superfluous for Coarse Task 1 but remain necessary for Coarse Task 2. (c) Relative performance of the fine-trained vs. coarse-trained models for three coarse tasks and different dataset sizes. The legend indicates the supposed level of boundary redundancy for each task. Models are one-hidden-layer networks with 30 hidden neurons,  trained with SGD. Averages are taken over 30 runs with different random initializations.  The test set consists of $10^4$ points, and the error bars represent the standard error.    }
    \label{fig:3}
\end{figure*}
 
The transition to a regime in which fine-grained training is no longer advantageous is consistently observed across all four real-world datasets analyzed (Fig.~\ref{fig:2}c). Moreover, a similar trend emerges when comparing architectures with varying numbers of hidden neurons (Fig.~\ref{fig:2}d): fine-grained training leads to higher accuracy for networks with more parameters, whereas coarse-grained training is preferable for models with fewer degrees of freedom.
 
{These results are not limited to image classification: the  same transition in relative performance can be observed  for a non-vision task. Specifically, we use as an illustrative example  a task of predicting forest cover type from a set of cartographic variables (Supplementary Materials, Fig.~S3; see the Methods section for details on the dataset). }

\subsection{The role of the label hierarchical structure,  data geometry, and the fine-coarse task alignment
}\label{sec:br}

The combination of these two observations suggests that the degree of over-parameterization is a crucial factor in selecting the best training strategy. Figure~\ref{fig:2}e shows the advantage of training on fine-grained labels as a function of the parameter-to-data ratio $n/p$, where $n$ is the total number of learnable parameters and $p$ is the dataset size. We observe a clear trend: fine-grained models tend to perform better at higher levels of over-parameterization ({namely}, larger $n/p$ values). However, the transition point appears to be dataset-specific, and the large fluctuations suggest the presence of additional relevant control variables.

Training on fine-grained labels does not always boost model performance and seems generally helpful when data is scarce compared to the number of parameters.  However, the role of the specific label structure is still an open question. The categories of a given dataset can be aggregated into different coarse classes, depending on the task. For a dataset like CIFAR-10, there is an inherent semantic organization of labels. For example, a natural choice would be to separate animals (divided into the fine classes of \textit{cats} {and} \textit{dogs}) from vehicles (\textit{cars} and \textit{trucks}). However, different coarse tasks could be defined by alternative label organizations; for instance, one could compose a coarse class with cats and trucks, and another with dogs and cars (Fig.~\ref{fig:3}a). Other datasets, such as MNIST or K-MNIST, do not present a natural organization and could be equally separated, for example, into even and odd numbers or numbers below and above 5. 
 
This section explores how different groupings of the same fine subclasses impact the relative performance of coarse- and fine-grained training. It is possible that training on fine labels adds information about the coarse classes and which features are relevant in discriminating them. Then, the specific class organization should be crucial in defining the amount of this additional information and its relevance for the task. 
 
The hypothesis we want to test is based on the concept of \textbf{boundary redundancy}, which intuitively captures the level of alignment  between the {decision boundaries required for the } tasks defined by the coarse and fine labels. Consider a coarse classification task,  such as separating animals from vehicles. In this case, some features, such as the presence of wheels or paws,  may allow the model to solve the coarse task perfectly, without necessarily learning to differentiate between fine-grained subclasses,  such as cats and dogs or trucks and cars (Fig.~\ref{fig:3}a). Instead, a model trained on fine-grained labels must learn distinctions that are not necessarily relevant for the coarse task.  
 In other words, the boundaries in the data space that the fine-trained model has to discover are \textit{redundant}  with respect to the coarse task on which performance is evaluated. This is the case illustrated on the left of Fig~\ref{fig:3}a and b.    
 
 In contrast, coarse classes could be composed of more granular categories whose discriminative features coincide with the ones needed for the coarse task (Fig.~\ref{fig:3}a, right panel). From a geometric point of view, the boundaries in the data space resulting from the two optimization strategies largely overlap; {namely}, redundancy is low.
 
Therefore, for class organizations with \textit{high boundary redundancy}, training on fine-grained labels may burden the optimization process with distinctions that are irrelevant to the task, potentially leading to equal or even worse performance. 
In contrast,  for class structures with  \textit{low boundary redundancy}, we hypothesize that fine-grained training is more likely to outperform its coarse counterpart, as the learning process focuses on the relevant decision boundaries while benefiting from additional structural information.
 
\begin{figure*}[t]
    \centering
    \includegraphics[width=0.87\linewidth]{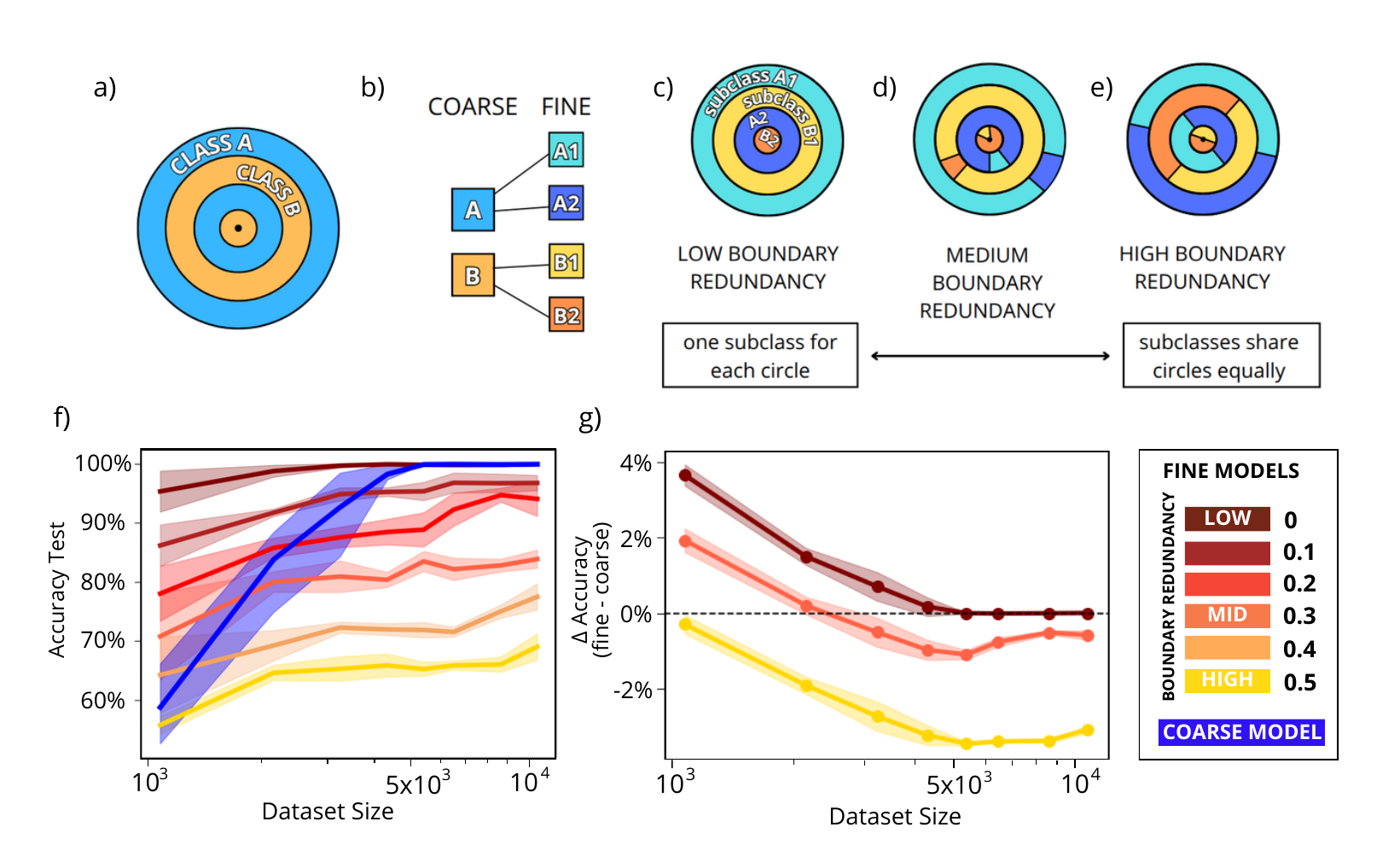}
    \caption {\textbf{Experiments with synthetic data confirm the role of data geometry for the advantage of fine-grained training.} (a) Dataset construction: four concentric circles ($K=4$), with randomly distributed points. Coarse labels alternate between class A (blue) and class B (orange) from one circle to the next.  (b) Hierarchical label structure obtained by splitting the two coarse classes into two subclasses. (c-e) The geometry of the fine classes can define problems with different boundary redundancies. (c)  No redundancy: each circle only contains elements of one subclass, and the fine-grained task and the coarse one coincide.  (d,e) By changing the fraction of points in each circle belonging to another subclass, the fine task diverges from the coarse one. The highest redundancy is obtained when subclasses are equally represented in each circle. (f-g) Performance of fine- vs. coarse-grained models trained on the synthetic circle dataset with $K=8$ as a function of dataset size ($N$), and different values of BR. Models trained with fine subclass structures of varying BR are colored in different shades of red, while the blue line represents the model trained on coarse labels. To equalize capacity, both models were matched for trainable parameters: the fine model uses 60 hidden neurons, the coarse model 150. Both models were trained with Adam (lr$=0.001$, $\beta_1=0.9$, $\beta_2=0.999$, $\epsilon=10^{-7}$). The training set consists of 5,000 points, and the test set consists of 3,000 points. In (f), the error bars indicate the range between the first and third quartiles of the data obtained on 30 runs with different initializations, while in (g), the error bars are the expected standard error.}
    \label{fig:circle_confs}
\end{figure*}
 
Fig.~\ref{fig:3}a provides three illustrative examples of class organizations with high, medium, and low boundary redundancy, constructed using the CIFAR-10 dataset. 
A two-dimensional t-SNE embedding of the data~\cite{maaten2008visualizing}  supports the geometric intuition behind the different levels of boundary redundancy (Fig.~\ref{fig:3}b).

To test our hypothesis, we trained two neural networks on fine- and coarse-grained groupings with different levels of boundary redundancy and compared their performance at different dataset sizes. The results are shown in Fig.~\ref{fig:3}c, where we consider the gain in performance $\Delta\text{Accuracy} = \text{Acc}_{\text{fine}}-\text{Acc}_{\text{coarse}}$ that training on fine-grained labels can provide relative to directly training the model on coarse labels.
For a task expected to have high boundary redundancy, the fine-grained model indeed never outperforms the coarse-grained one. In contrast, the fine-grained model consistently generalizes better on a task with low boundary redundancy. At an intermediate level of boundary redundancy, we observe the transition described in the previous section: the fine-trained model performs better for smaller dataset sizes.

Although this result fully supports our hypothesis, the notion of boundary redundancy is mainly intuition-based and difficult to quantify. 
{We could not find a simple measurable definition of boundary redundancy that generalizes to all possible data structures. 
However, we introduce two simple proxies that are intuitively related and significantly correlate with the improvement from coarse- to fine-grained training. Namely, we consider  the accuracy of a linear classifier (logistic regression) on the coarse task, and the average  silhouette score of the coarse classes~\cite{rousseeuw1987silhouettes} (a measure of how close a data point is to points of the same class compared to points in the other class). 
Figure~S4 of the Supplementary Materials shows the  negative correlation between both quantities and the difference in accuracy between the fine- and coarse-grained models,  when trained on different geometric data configurations. Specifically, we grouped the eight classes of FashionMNIST into different pairs of coarse classes. When the coarse task is relatively \textit{simple} or  nearly linearly separable, the boundary redundancy is expected to be high, and training on fine-grained classes brings little to no advantage.  Although these quantities do not provide a formal definition of boundary redundancy, they demonstrate that meaningful geometric proxies can be defined and used in practice.

The next section introduces a synthetic dataset in which redundancy can be directly measured and controlled,  allowing a more stringent quantitative test of the relationship between data geometry and the advantage of fine-grained training.}
 
\subsection{Testing the importance of boundary redundancy using a controlled synthetic dataset} \label{sec:circles}
 
To quantitatively control the level of boundary redundancy inherent in a label structure, we introduce a synthetic dataset based on a radial function. 
These kinds of functions are known to be difficult for shallow neural networks to learn~\cite{pmlr-v49-eldan16}, making them a suitable choice for constructing a sufficiently complex task. At the same time, their two-dimensional nature allows for straightforward visualization and interpretation. By adjusting the geometry of the data points across subclasses, we can continuously tune the degree of redundancy. 
More specifically, the two-dimensional dataset consists of data points $(x_1,x_2)$ arranged in $K$ concentric circles.  The two coarse labels are assigned to points on a given circle, alternating between class A and class B as the circle radius increases (Fig.~\ref{fig:circle_confs}a). Thus, the coarse class of a point $x^\mu = (x_1^\mu,x_2^\mu)$ is given by
 
\begin{equation}
\begin{cases}
    Y^\mu = 0 \quad \text{if}\; \| x^\mu\| = 2i/K, \\
    Y^\mu = 1 \quad \text{otherwise} 
\end{cases}\;i = 1,\dots,K/2, 
\end{equation}
 where the label 0 corresponds to class A, and the label 1 to class B. 
 
Each one of these coarse classes is then partitioned into $K/2$ subclasses (\ref{fig:circle_confs}b), and we engineered this partitioning to create structures with different levels of redundancy. 
The simplest way to define the subclasses is to assign each circle to a distinct subclass, as illustrated in Fig.~\ref{fig:circle_confs}c:
 
 \begin{equation}
     y^\mu = j \quad \text{if } \left\| x^\mu \right\| = r_j, \quad j = 1, \dots, K, 
 \end{equation}
 
 where $r_j$ is the radius of the $j$-th circle. 
 This configuration represents the structure with the lowest level of boundary redundancy. To separate the fine classes ({namely}, individual circles), the model has to identify the same boundaries needed to separate the two coarse classes. 
 
 To tune the level of boundary redundancy, we can redefine the subclasses by mixing data regions belonging to different circles (Fig.~\ref{fig:circle_confs}). The level of mixing can be quantified by measuring the fraction of each circle that is occupied by a subclass different from the predominant one. This fraction determines the extent to which the task defined by fine and coarse training overlaps.  When each circle defines a distinct subclass, the boundary redundancy is zero (Fig.~\ref{fig:circle_confs}c).
 Figure~\ref{fig:circle_confs}d shows an intermediate case in which each circle contains 70\% of points assigned to a dominant subclass and the remaining 30\% to another subclass, thus corresponding to a redundancy level of $0.3$.
 The highest possible redundancy is $1 - 2/K$, as represented in Fig.~\ref{fig:circle_confs}e. 
 All configurations share the same coarse label structure: even when a single circle contains points from multiple subclasses, all points on that circle still belong to the same coarse class, following the hierarchical organization illustrated in Fig.~\ref{fig:circle_confs}b.
 
Figure~\ref{fig:circle_confs}f shows the test accuracy on the coarse classification task for models trained on fine-grained labels with varying levels of boundary redundancy. As reported in the previous section, fine-trained models with a fixed architecture outperform the coarse-trained model (blue curve) only for small dataset sizes. As we hypothesized, the accuracy achieved by fine-trained models scales with the boundary redundancy of the label structure. As the dataset size increases, the benefit of fine-grained supervision diminishes (Fig.~\ref{fig:circle_confs}g), and only label structures with low boundary redundancy continue to offer an advantage in training for larger datasets. The transition point, {namely} the dataset size at which training on fine labels is preferable, crucially depends on the data geometry, in particular on the boundary redundancy. 
When there is no boundary redundancy (brown curve), the problem of separating the fine classes is geometrically equivalent to the coarse class discrimination. Therefore, the model trained on fine classes never performs worse than the coarse-trained one. It can only do better when data is scarce. 
 
 Compared to the real-world datasets discussed in the previous section,  a key difference is that this synthetic task has negligible noise. In other words, generalization and training accuracy typically coincide in this setting. Therefore, when the data are insufficient for the coarse model to learn effectively, training on fine-grained subclasses provides additional information to the optimization process, helping convergence towards the correct boundaries. 
 
\subsection{Insights on the relation between  fine and  coarse training from a loss function decomposition } \label{sec:losses}
We consider a setting in which there are $K$ fine-grained classes, denoted by the labels  $y_i \in \{0, 1\}$, with $i = 1, \dots, K$. Each data point belongs to exactly one fine-grained class, so that $\sum_{i=1}^{K} y_i^\mu = 1$, where the index $\mu$ identifies the data point. 
 
We introduce two coarse disjoint  classes with index sets $\mathcal{C}_0$ and $\mathcal{C}_1$, such that $\mathcal{C}_0 \cup \mathcal{C}_1 = \{1, \dots, K\}$ and $\mathcal{C}_0 \cap \mathcal{C}_1 = \emptyset.$
The corresponding coarse labels $Y^\mu \in \{0, 1\}$ are defined as $Y^\mu = \sum_{i \in \mathcal{C}_0} y_i^\mu$.

The fine-trained model outputs a probability distribution $\hat{y}_i^\mu$ over the fine classes using a softmax function. The predictions over the coarse classes $\hat Y^\mu$ are simply obtained by summation, given the known label structure, namely

\begin{equation}
\hat Y^\mu = \sum_{i \in \mathcal{C}_0} \hat y_i^\mu. 
\end{equation}

We can then formulate the loss for  coarse training as a classic  binary cross-entropy  
 
\begin{gather}
        \mathcal{L}_{\text{coarse}}= \nonumber\\ =- \frac{1}{p} \sum_{\mu=1}^p \left( Y^\mu \log(\hat{Y}^\mu) + (1-Y^\mu) \log(1-\hat{Y}^\mu)\right)=\nonumber\\
         =- \frac{1}{p} \sum_{\mu=1}^p  \Bigg [\Big(\sum_{i \in \mathcal{C}_0}  y_i^\mu\Big)\log\Big(\sum_{i \in \mathcal{C}_0} \hat y_i^\mu\Big)\,+ \\+\,\Big(\sum_{i \in \mathcal{C}_1} y_i^\mu\Big)\log\Big(\sum_{i \in \mathcal{C}_1} \hat y_i^\mu\Big)\Bigg ].\nonumber
\end{gather}
 
Analogously, the loss $\mathcal{L}_{\text{fine}}$  that is minimized during fine{-}training is a cross-entropy loss  over the $K$ classes, defined as 
 
\begin{equation}
        \mathcal{L}_{\text{fine}} = \frac{1}{p} \sum_{\mu=1}^p \sum_{i \in \mathcal{C}_0\cup\,\mathcal{C}_1} y_i^\mu \log(\hat y_i^\mu).
\end{equation}

However, this expression can be reformulated  as  
 
\begin{equation}
        \mathcal{L}_{\text{fine}} = \mathcal{L}_{\text{coarse}} +\mathcal{L}_{\text{intra-class}}, 
\end{equation}
 
where the intra-class loss is
 
\begin{gather}
        \mathcal{L}_{\text{intra-class}} =\nonumber\\ \frac{1}{p} \sum_{\mu=1}^p \Bigg[\sum_{i \in \mathcal{C}_0} y_i^\mu \log\Big(1+\dfrac{\sum\limits_{j \in \mathcal{C}_0\setminus \{i\}}\hat y_j^\mu}{\hat y_i^\mu}\Big)+ \\ + \sum_{i \in \mathcal{C}_1} y_i^\mu \log\Big(1+\frac{\sum\limits_{j \in \mathcal{C}_1\setminus \{i\}}\hat y_j^\mu}{\hat y_i^\mu}\Big)\Bigg].\nonumber 
\end{gather}

\begin{figure*}[t]
    \centering
    \includegraphics[width=0.75\linewidth]{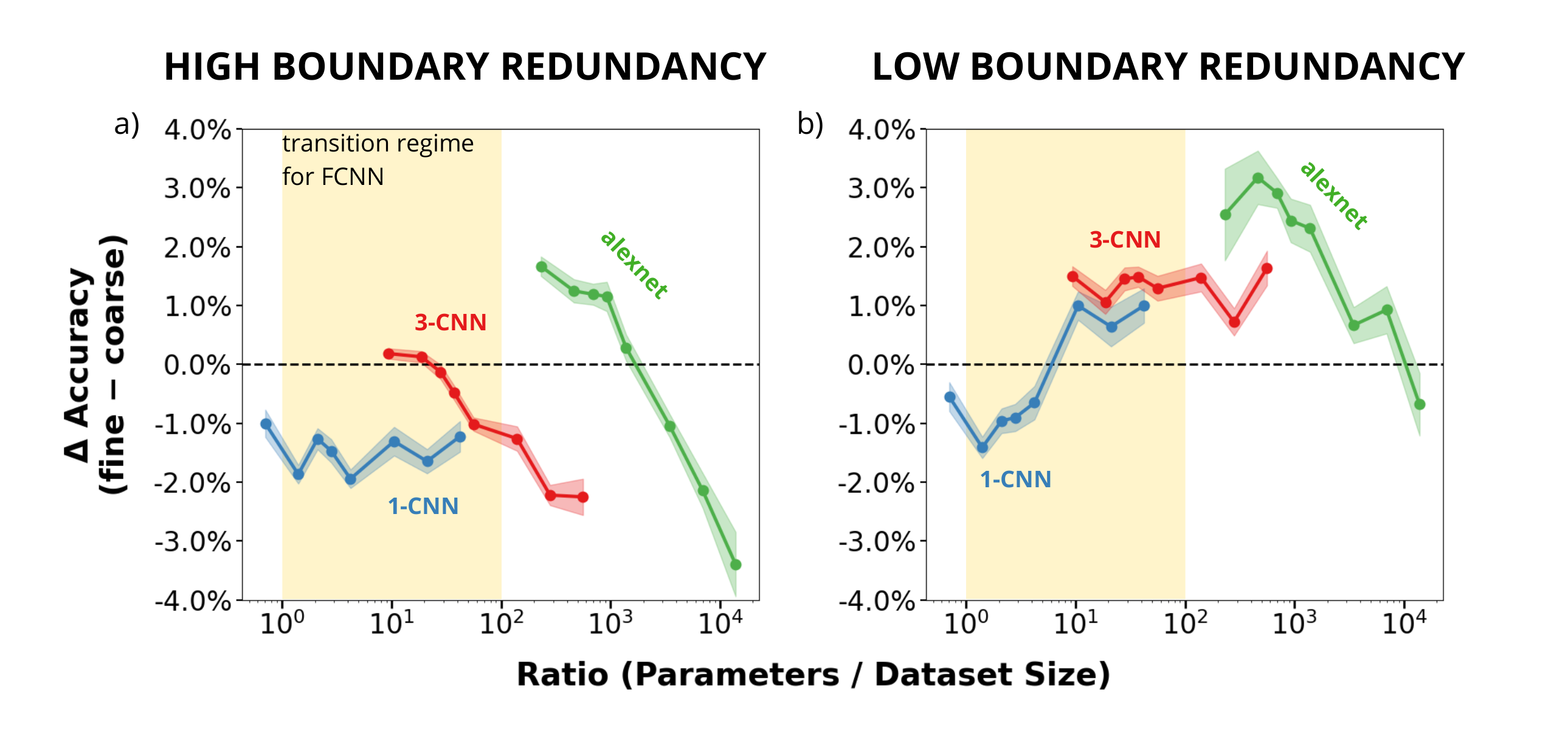}
    \caption{{\textbf{The advantage of training convolutional neural networks with fine-grained labels depends on  boundary redundancy and model complexity.} $\Delta\text{Accuracy} = \text{Acc}_{\text{fine}} - \text{Acc}_{\text{coarse}}$ is reported as a function of the ratio between the number of parameters of the coarse-grained model and the training set size, for three convolutional networks with one (1-CNN, blue) and  three (3-CNN, red) convolutional layers and for the classic AlexNet architecture (alexnet, green). (a) Coarse task with high boundary redundancy, obtained by grouping CIFAR-10 classes into vehicles (airplane, automobile, ship, truck) vs. animals (dog, deer, bird, cat). (b) Coarse task with low boundary redundancy, obtained by grouping (airplane, dog, ship, deer) vs (truck, cat, car, bird). All networks were trained with SGD (momentum 0.9), a linearly decreasing learning rate, and a batch size that scales with dataset size. Dataset size is varied by changing the number of data points from 400 to 24000. The yellow band marks the transition regime found for the fully connected networks of Fig.~\ref{fig:2}e. Error bars are the standard error over 10 runs with different initializations.}}
    \label{fig:cnn_transition}
\end{figure*}
 
This decomposition shows that training on fine-grained classes minimizes the coarse loss together with a term that depends on intra-class errors, namely, misclassifications of fine labels within a coarse class. If the boundaries that optimize the coarse and fine tasks largely coincide, minimizing the intra-class loss does not add substantial complexity to the optimization. Therefore, in this case, fine-training can improve predictions on the coarse-grained task.
This is particularly true when the training set is small, as the additional term $ \mathcal{L}_{\text{intra-class}} $ can effectively act as a regularizer, pushing the model - among all solutions that minimize the coarse loss - towards one that also separates subclasses reasonably well. This, in turn, can encourage better representations and improve generalization.
 
Conversely, when the boundaries for the two tasks differ substantially, minimization with the additional intra-class loss term can select solutions that make more errors on the coarse classes. This effect is particularly relevant for large training sets relative to the number of free parameters, for which the optimization is already strongly constrained.

\subsection{Extension to Convolutional Neural Networks}

While the goal of this work is to focus on simple models  to extract interpretable principles, a natural question is how much our findings extend to different architectures. Along this direction, we tested the advantage of fine-grained training using Convolutional Neural Networks (CNNs), a widely used architecture for computer vision tasks.  
Figure \ref{fig:cnn_transition} shows the results obtained training two basic architectures, with one convolutional layer or with three convolutional layers, as well as a deeper state-of-the-art model, AlexNet\cite{Krizhevsky2012}. The models are trained on two different configurations of eight classes of CIFAR10. The two alternative groupings of these eight classes intuitively represent coarse tasks with high and low boundary redundancy. 
The high-boundary-redundancy configuration groups semantically related subclasses into vehicles (airplane, automobile, ship, truck) and animals (dog, deer, bird, cat). The low-boundary-redundancy configuration is instead  obtained by grouping visually dissimilar subclasses (caption of Fig.~\ref{fig:cnn_transition}), likely resulting in a substantially higher alignment between the fine- and coarse-grained decision boundaries.  
The first result is that the advantage of training on fine-grained labels is greater for the configuration with lower boundary redundancy for all different architectures. This is clear by comparing the advantage relative to single architectures in panels a and b, confirming the importance  of data geometry in setting the preferred strategy across different architectures.

The behavior of the 1-CNN model (blue curve), which has a ratio of parameters-to-dataset-size close to that of the fully connected networks in the previous sections,  perfectly aligns with our previous findings. Indeed, training on fine-grained classes improves performance as we increase the overparametrization. 
However, for more complex models, such as AlexNet, operating in a highly overparameterized regime, an interesting inverse transition is observed, suggesting  that additional mechanisms become relevant. 
Therefore, the role of boundary redundancy extends beyond fully connected networks and remains relevant for convolutional architectures. Similarly,  the transition between regimes in which fine- or coarse-grained training is preferable appears to be a general phenomenon rather than a consequence of the specific architectural choice.  However, the largely over-parameterized regime explored by more complex models displays additional  phenomena that remain to be fully characterized.

\section{Discussion}

This work represents a first step toward understanding the role of hierarchical class structures in the training of neural networks. Such structures are ubiquitous: many standard categorical datasets are organized hierarchically, including classic computer vision benchmarks such as ImageNet~\cite{deng2009imagenet} or CIFAR-10 and CIFAR-100~\cite{krizhevsky2009learning}. However, as discussed in the introduction, such label structures are not limited to specific datasets, but rather reflect a general tendency to record and organize knowledge in progressively more detailed categories. The organization of Wikipedia pages is one example~\cite{suchecki2012evolution}. Words themselves of natural language are naturally arranged into hierarchies through structural and semantic relations, as exemplified by simple models of generative grammars ~\cite{degiuli2019random} and by the WordNet lexical database~\cite{miller1995wordnet}.

Empirical studies have observed that we can leverage these data hierarchical structures in machine learning, and, more specifically, that training on labels that are more fine-grained than the target task labels can improve performance~\cite{dermatologist, Chen, ridnik2021imagenet,hong2024}. However, our results show that the picture is more nuanced. We uncover a complex interplay between data geometry, label organization, and the degree of model overparameterization. In particular, fine-grained training can boost generalization when the available training data are scarce relative to the number of model parameters. In this regime, the additional term, appearing in the cross-entropy decomposition we uncovered, acts as a form of regularization, guiding the optimization toward solutions that also separate subclasses effectively. {Fine-training improves generalization, likely} because it promotes better data representations. 
 
This benefit, however, is task- and dataset-dependent. The performance gain depends on a trade-off between the additional optimization burden of fitting fine-grained labels and the degree of alignment between the coarse and fine decision boundaries. Our phenomenological study highlights the subtleties of the competing effects of injecting additional information via subclasses and the optimization cost that training on them entails. 
 
We have addressed the problem in a controlled setting,  using standard datasets,  low-dimensional synthetic data, and training relatively small, shallow fully-connected neural networks with standard optimization techniques. This \textit{minimalistic} approach, inspired by statistical physics, allows an extensive quantitative phenomenological exploration and transparent interpretation of the results. However, we also started to test our findings across a wider range of architectures   and datasets to assess their generality. In particular, we considered   convolutional neural networks. In this case as well, the advantage of fine-training depends on the level of boundary redundancy, and the optimal  training strategy changes with  dataset size and  model capacity. However, experiments with deep and complex models,  such as AlexNet, also show an additional transition in the strongly overparametrized regime, suggesting a richer scenario yet to be fully explored. This opens up several avenues for future research, including a systematic investigation of how different architectures, such as deeper convolutional networks (for example, ResNet) and attention-based transformers, influence the interplay between hierarchical label structures, data geometry and learning.

The presence of a hierarchical label structure often reflects an underlying compositionality in the features defining the classes: general features distinguish coarse categories, while increasingly specific features are required to resolve finer distinctions. Compositionality is widely regarded as a core ingredient of neural network generalization properties~\cite{poggio2020theoretical,danhofer2025position,lin2017does}. In this perspective, our results suggest that understanding the dataset-specific relationship between label and data structure can be relevant for designing training strategies that leverage compositionality effectively.
Moreover, data could also be inherently compositional and naturally organized in hierarchical structures,  even when the labels attached to them do not reflect this organization.
This hidden stratification can be inferred and used to benefit training~\cite{sohoni2020no}. Therefore, we expect our approach to be extendable to unsupervised learning settings, particularly when handling compositional data.

While our focus has been on identifying basic principles underlying the advantage of fine-grained training, a natural next step is to leverage these principles to optimize neural architectures and training strategies. Several previous works have indeed attempted to incorporate hierarchical class structures directly into models.
For example, hierarchical loss functions reflecting the label structure have been explored for standard neural network architectures~\cite{ wu2019hierarchical}. However, such approaches often failed to outperform conventional cross-entropy training, except in cases with very limited training data. Another line of work sought to encode the hierarchy directly into the architecture itself, for instance, by separating the classification of labels at different levels using specific layers with dedicated cross-entropy losses~\cite{yan2015hd,la2021learn}. Although these methods showed marginal performance improvements, they require substantial customization and implementation complexity,  and they did not become common practice.
 
A deeper understanding of the interplay between data characteristics,  label structure, and optimization may help guide future developments in this direction. As a simple example, the loss decomposition in Section~\ref{sec:losses} suggests an alternative training objective to test:  
$\mathcal{L} = \mathcal{L}_{\text{coarse}} + \beta\mathcal{L}_{\text{intra-class}}$,  with $\beta\leq1$. A dataset-specific tuning of the hyperparameter  $\beta$ could be a possible way to control the trade-off between the advantage given by the subclasses' information and the increased complexity of the optimization.

More broadly, our work relates to a fundamental question in machine learning: given a classification task, is it preferable to train a specialized model with many examples labeled at the target resolution, or to first learn from a richer, more general supervision signal and then adapt to the task of interest? Training on fine-grained labels can be viewed as a form of pre-training, but instead of learning from a different dataset, the model is exposed to a more detailed label structure within the same data.
 
The great success of transfer learning~\cite{yosinski2014transferable} and, more recently, of foundation models in natural language processing~\cite{radford2018improving,bommasani2021opportunities} suggests that large-scale, non-specific pre-training is generally advantageous. However, as in the case of transfer learning, the gains can depend critically on data properties and task alignment~\cite{zamir2018taskonomy,kornblith2019better,achille2021information},  {as well as on the interplay between intra- and inter-class diversity~\cite{zhang2023trade}}.
{This is also a classic tension in multitask learning where there is a trade-off between the helpful additional information provided by  auxiliary labels and the possible interference between tasks, making optimization more difficult~\cite{vandenhende2021multi}}. 
Our findings similarly reveal that the benefits of fine-grained training are context-dependent. 
Interestingly, recent theoretical work has begun to address the role of geometrical properties of data manifolds~\cite{goldt2020modeling} and hierarchical data composition~\cite{cagnetta2024deep} in neural network learning. Combining our quantitative, phenomenological description with these approaches from statistical physics could offer a path toward a precise characterization of the relationship between network training and label granularity, ultimately providing actionable and explainable guidelines for practitioners.

\section{Methods}\label{methods}
 
\subsection{Datasets used}\label{methods:datasets}
 
We considered for five analysis the following  widely used  datasets for classification:
\begin{itemize}
\item MNIST, handwritten digits, 28x28 greyscale images~\cite{lecun1998gradient},
\item Kuzushiji-MNIST, or K-MNIST;
cursive Japanese characters, 28x28 greyscale images~\cite{clanuwat2018deep},
\item 
Fashion MNIST, or F-MNIST;
Zalando's article images,
28x28 greyscale images~\cite{xiao2017fashion},
\item CIFAR-10,
32x32 RGB images~\cite{krizhevsky2009learning},
\item {Covertype~\cite{covertype_31}, forest cover type of 30 x 30 meter cells from 54 cartographic variables only. }
\end{itemize}

These datasets are organized into classes, such as handwritten digits in the MNIST dataset or types of vehicles and animals in CIFAR-10. 
We created hierarchical label structures by aggregating these classes into two coarse-grained, larger classes in different ways. For example,  digits from 0 to 7 can be grouped into the two larger classes corresponding to even and odd numbers. All image pixels are normalized to the range $[0,1]${, while non-image datasets are standardized so that each feature has zero mean and unit standard deviation}.

\subsection{Model architecture}\label{methods:models}
 
Given a dataset $\mathcal{D} = \{x^\mu, y^\mu\}_{\mu=1}^P$, where $x^\mu \in \mathbb{R}^{d}$ are feature vectors and $y^\mu \in \mathbb{R}^K$ are one-hot encoded fine-grained labels over $K$ subclasses, the fine model is a fully connected neural network with $K$ outputs and one hidden layer composed of $N_{\text{fine}}$ hidden neurons. The following equations define the network function: 
 
\begin{equation}
\begin{split}
\hat y_k^\mu = & \text{softmax}_k\big(u(x^\mu)\big),\\
u_k(x^\mu) = & \sum_{i=1}^{N_{\text{fine}}} v^k_i \,\text{ReLU}(\omega_i \cdot x^\mu + b_i) + \beta_k, 
\end{split}
\end{equation}
with $\omega_i \in \mathbb{R}^{d}$, $b_i,\,v_i^k,\, \beta_k \in \mathbb{R}$. The results we present are not dependent on the choice of activation function; we obtained similar results with the tanh function, as shown in the Supplementary Materials, Fig.~S{5}.
The model is trained with stochastic gradient descent and a standard cross-entropy loss.
 
The coarse model (Fig.~\ref{fig:dataset_subdivision}b) is instead a binary classifier with a sigmoid function as output:
 
\begin{equation}
\hat Y^\mu_{\text{c}} = \sigma\left(\sum_{i=1}^{N_{\text{coarse}}} v_i \,\text{ReLU}(\omega_i \cdot x^\mu + b_i) + \beta\right).
\end{equation}
 
 $N_{\text{coarse}}$ is the number of hidden neurons, $\sigma(\cdot)$ is the sigmoid activation function, and $\omega_i \in \mathbb{R}^{d}$, $b_i,\,v_i,\, \beta \in \mathbb{R}$. The model is trained with a binary cross-entropy loss defined using the coarse labels $Y^\mu = \sum_{k\in \mathcal{C}_0} y_k^\mu$, with $\mathcal{C}_0$ and $\mathcal{C}_1$ indicating the two coarse classes. All model weights are initialized with Glorot uniform initialization, and the biases are set to zero.
 
 We are interested in comparing the performance of these two models on the coarse-grained task. 
 The prediction of the fine-grained model on the coarse labels is obtained by summing the predicted subclass probabilities (Fig.~\ref{fig:dataset_subdivision}c). 
 The test accuracy for both models is the proportion of correctly predicted  coarse labels in the test set:
 
\begin{equation}
\begin{split}
\text{Acc}_{\text{coarse}} &= \frac{1}{P} \sum_{\mu=1}^P \mathbb{I}\left(\Theta(\hat Y_{\text{c}}^\mu-\frac{1}{2}) = Y^\mu\right), \\
\text{Acc}_{\text{fine}} &= \frac{1}{P} \sum_{\mu=1}^P \mathbb{I}\left(\Theta(\hat Y_{\text{f}}^\mu -\frac{1}{2})= Y^\mu\right),
\end{split}
\end{equation}
 
with $\hat Y_{\text{f}}^\mu = \sum_{k\in \mathcal{C}_0} \hat y_k^\mu$ representing the coarse prediction obtained by summing the softmax outputs over the corresponding coarse class, and $\Theta(\cdot)$ is the Heaviside step function. Training accuracy also refers to the proportion of correctly predicted coarse labels in the training set.
To ensure a fair comparison between the two networks in Fig.~\ref{fig:dataset_subdivision}, we conducted extensive experiments, tuning the number of hidden neurons, $N_{\text{coarse}}$ and $N_{\text{fine}}$, so that both architectures had roughly the same number of learnable parameters. Notably, the results do not depend on this precise architectural tuning.

{\paragraph{Convolutional networks.} In Fig. \ref{fig:cnn_transition}, we introduce three convolutional architectures of increasing depth and number of parameters:
\begin{itemize}
\item 1-CNN: one convolutional layer (32 filters, $3\times3$, ReLU); $2\times2$ max pooling; dense layer of 64 units before the output.
\item 3-CNN: three convolutional layers (32, 64, and 128 filters, $3\times3$, ReLU); $2\times2$ max pooling; dense layer of 64 units before the output.
\item AlexNet-like architecture \cite{Krizhevsky2012}; we adapted the original architecture to  input images of CIFAR10 size.
\end{itemize}
As in Sec.~4.2, coarse- and fine-grained models share the same convolutional structure and differ only in the output layer (two units for the coarse task, eight for the fine task). Models are trained with SGD (momentum $0.9$), training is stopped at convergence, checking for a minimum change in the loss of  $5\times10^{-4}$ (with a patience of 30 epochs), and a linearly decaying learning rate. Batch size scales with dataset size, from 8 to 64. Each combination of architecture, dataset size, and coarse task is trained 10 times with different initializations.}
 
\subsection{t-SNE representation of CIFAR dataset}\label{methods:tsne}
 
To provide visual intuition for the concept of boundary redundancy introduced in Section~\ref{sec:br}, we trained a convolutional neural network on a subset of CIFAR-10 containing four classes: \textit{cats}, \textit{dogs}, \textit{cars}, and \textit{trucks}. We used 5000 training examples per class. The CNN architecture consisted of three convolutional layers with ReLU activations (32, 64, and 64 filters, respectively; kernel size 3×3; stride 1). The first two convolutional layers are followed by 2×2 max-pooling layers (stride 2). The convolutional stack was followed by two dense layers: one with 64 neurons and ReLU activation, and an output layer with four neurons and softmax activation. The model was trained for 30 epochs using the Adam optimizer (learning rate $10^{-3}$), sparse categorical cross-entropy loss, and a batch size of 64.
 
After training, we extracted the last hidden layer representation (the 64-dimensional layer before the softmax) for a subset of the training set of 500 examples from each class. These high-dimensional representations capture the features that the network has learned to distinguish between the four classes.
 
We then applied t-SNE dimensionality reduction with two components and 600 iterations to visualize these internal representations in a 2D space. While t-SNE may introduce visualization artifacts and does not preserve exact geometric relationships, the qualitative patterns provide intuitive support for our concept of boundary redundancy.

\subsection{Coding availability}
 
The code for the results in this article is available at:
 
\href{https://github.com/daviiiiide/subclasses}{https://github.com/daviiiiide/subclasses}
 
\section{Author Contributions}
 
MO conceived and supervised the project. DP and FM   conducted the numerical experiments and analyzed the results. MO, DP, and FM wrote the manuscript. 
MC and PF assisted with supervision and funding. All authors reviewed the manuscript. 
 
\section{Acknowledgements}
 
We gratefully acknowledge Simone Sciandra and Claudio Caprioli for their insightful discussions and contributions to this work.
 
\section{Funding}
 
Funding: The Ph.D. fellowship of F. Milanesio is co-financed by the company Additati\&Partners Consulting s.r.l at UniTO.
 
\bibliographystyle{naturemag}
\bibliography{references} 
\onecolumn 
\begin{appendices}
\section*{Supplementary Material}
\addcontentsline{toc}{section}{Supplementary Material} 
\renewcommand{\thefigure}{S\arabic{figure}}
\setcounter{figure}{0}
\begin{figure}[ht]
    \centering
    \includegraphics[width=0.92\textwidth]{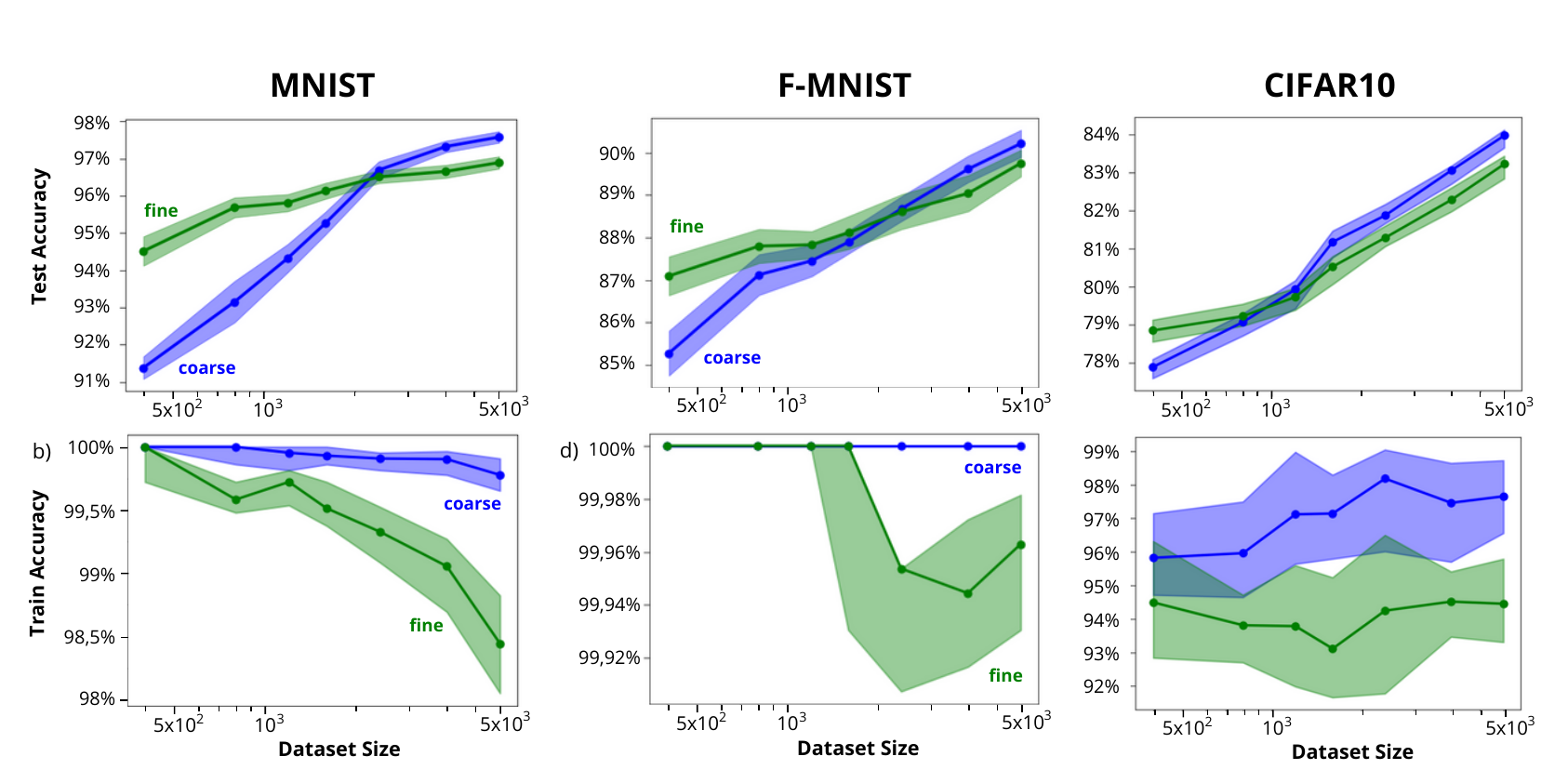}
    \caption{\textbf{Train and test accuracy of different datasets.} Test accuracy and corresponding train accuracy of models trained on fine-grained labels (green) and coarse-grained labels (blue) across subsets of various sizes for the MNIST, F-MNIST, and CIFAR10 datasets. Networks with 10 neurons, decreasing the learning rate from 0.01 to 0.001; the batch size was adjusted according to dataset size: a batch size of 8 was used for the smallest dataset, and 32 for the largest. SGD with early stopping was used as the optimizer. The test set consists of 104 points, and the error bars represent the range between the first and third quartiles of the data obtained from 30 runs with different weight initializations.}
\end{figure}

\begin{figure}[ht]
    \centering
    \includegraphics[width=0.9\textwidth]{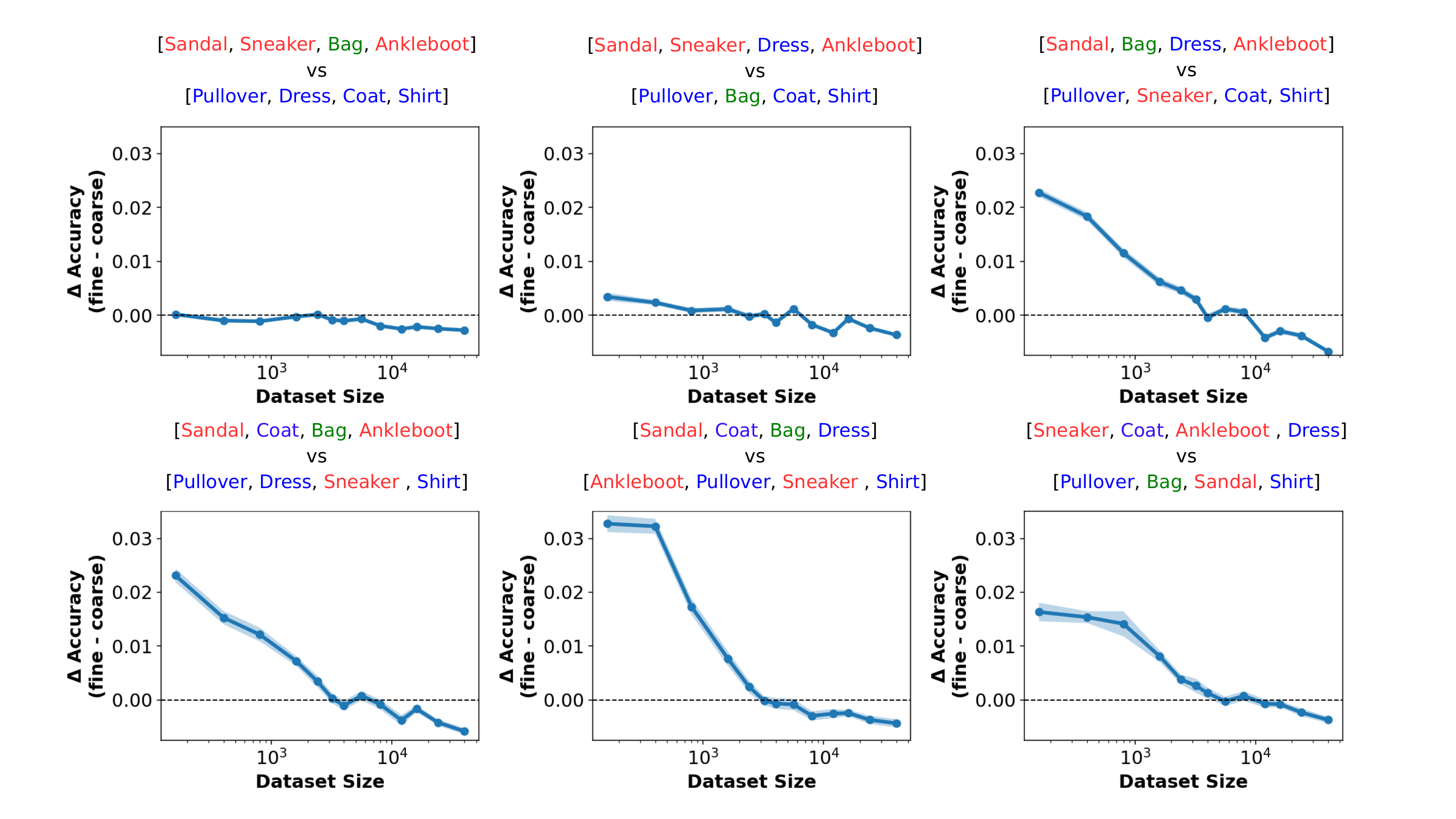}
    \caption{\textbf{Transition for different configurations of the F-MNIST dataset.} The difference in test accuracy, $\Delta \text{Accuracy} = \text{Acc}_{fine} - \text{Acc}_{coarse}$, across different configurations of the F-MNIST dataset at different dataset sizes. Networks with 10 hidden neurons trained with SGD and early stopping. The error bars represent the standard error over 30 runs. Different semantic groupings are colored differently to highlight the boundary redundancy of the different coarse labels.
}
\end{figure}

\begin{figure}[ht]
    \centering
    \includegraphics[width=0.45\textwidth]{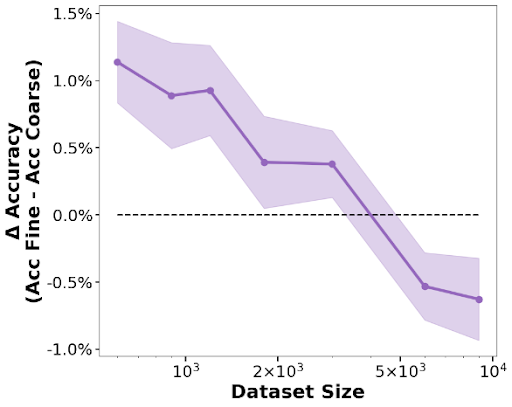}
    \caption{{\textbf{Transition in the advantage of fine-grained training on tabular data.} $\Delta \text{Accuracy} = \text{Acc}_{fine} - \text{Acc}_{coarse}$ is reported as a function of training set size for the Forest Cover Type dataset.  The least populated class was dropped, and the remaining six classes of cover types were split into two coarse classes of three subclasses each (i.e., Spruce, Ponderosa Pine, Krummolz vs Lodgepole Pine, Aspen, Douglas-Fir).  The model used is a one-hidden-layer neural network trained with SGD (momentum 0.9), a linearly decaying learning rate starting from 0.01, and trained to convergence  (a minimum loss variation of  0.0005 with patience of  30 epochs). Batch size scales with dataset size, from 8 to 64. The error bar represents the standard error over 20 independent runs.}
}
\end{figure}

\begin{figure}[ht]
    \centering
    \includegraphics[width=0.8\textwidth]{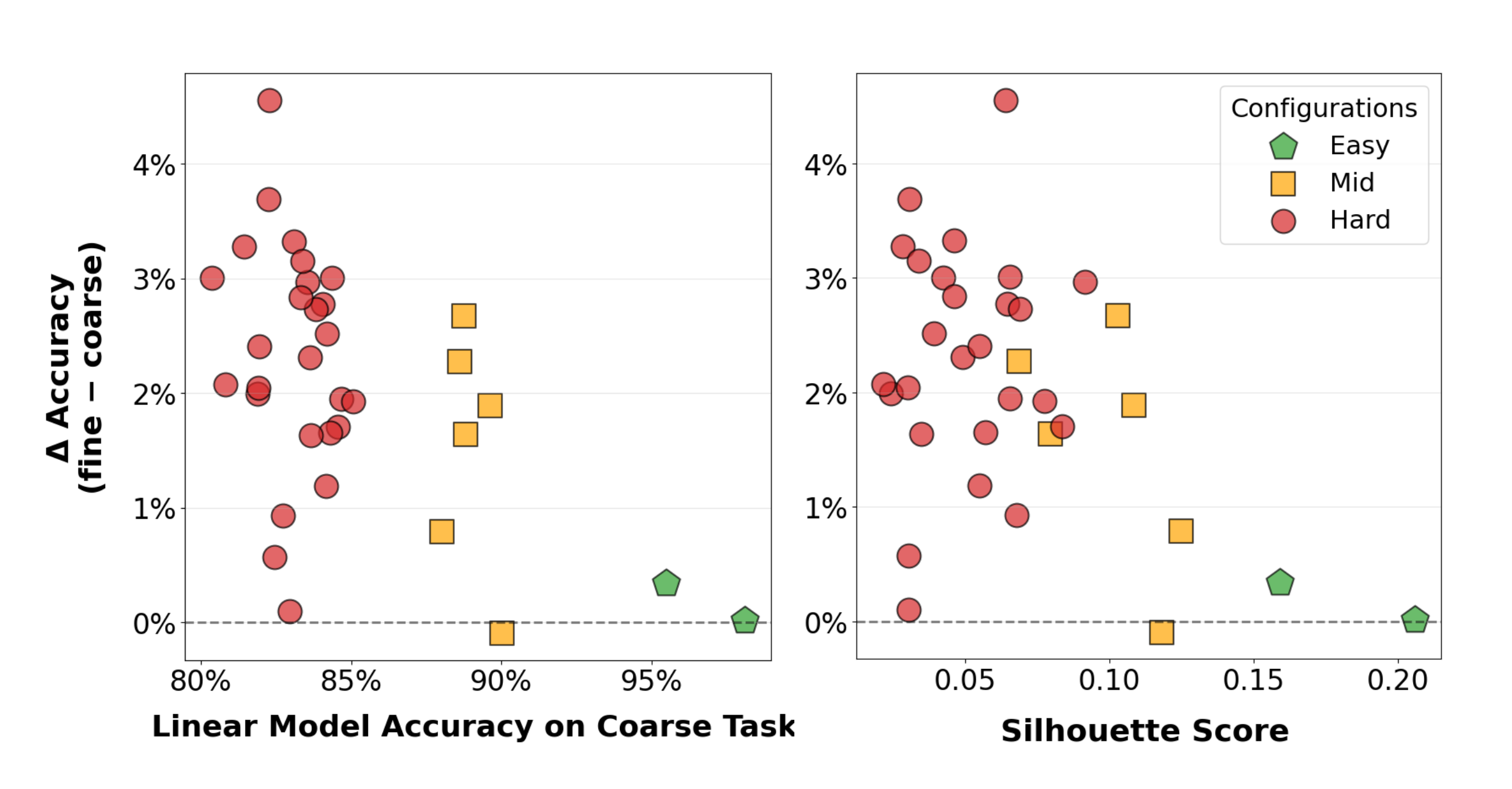}
    \caption{{\textbf{Geometric proxies for boundary redundancy.} Each point corresponds to a different grouping of the eight F-MNIST classes into two coarse categories, plotted against the resulting gain from fine-grained training,  $\Delta \text{Accuracy} = \text{Acc}_{fine} - \text{Acc}_{coarse}$. Two proxies were used: (left) the test accuracy of a logistic regression trained directly on the coarse labels, and (right) the silhouette score of the two coarse classes computed in input space. Configurations are colored as Easy, Mid, or Hard by applying k-nearest neighbors to the linear model accuracy, grouping configurations of comparable coarse-task difficulty. The two proxies are coherent with each other, since they rank configurations consistently. In fact, the clustering of configurations obtained with the linear model accuracy (symbols) is roughly reproduced by the silhouette score.  Both proxies are negatively correlated with  $\Delta \text{Accuracy}$ (Pearson $r = -0.62$ for the linear model, $r = -0.43$ for the silhouette score). }
}
\end{figure}

\begin{figure}[!t]
    \centering
    \includegraphics[width=0.85\textwidth]{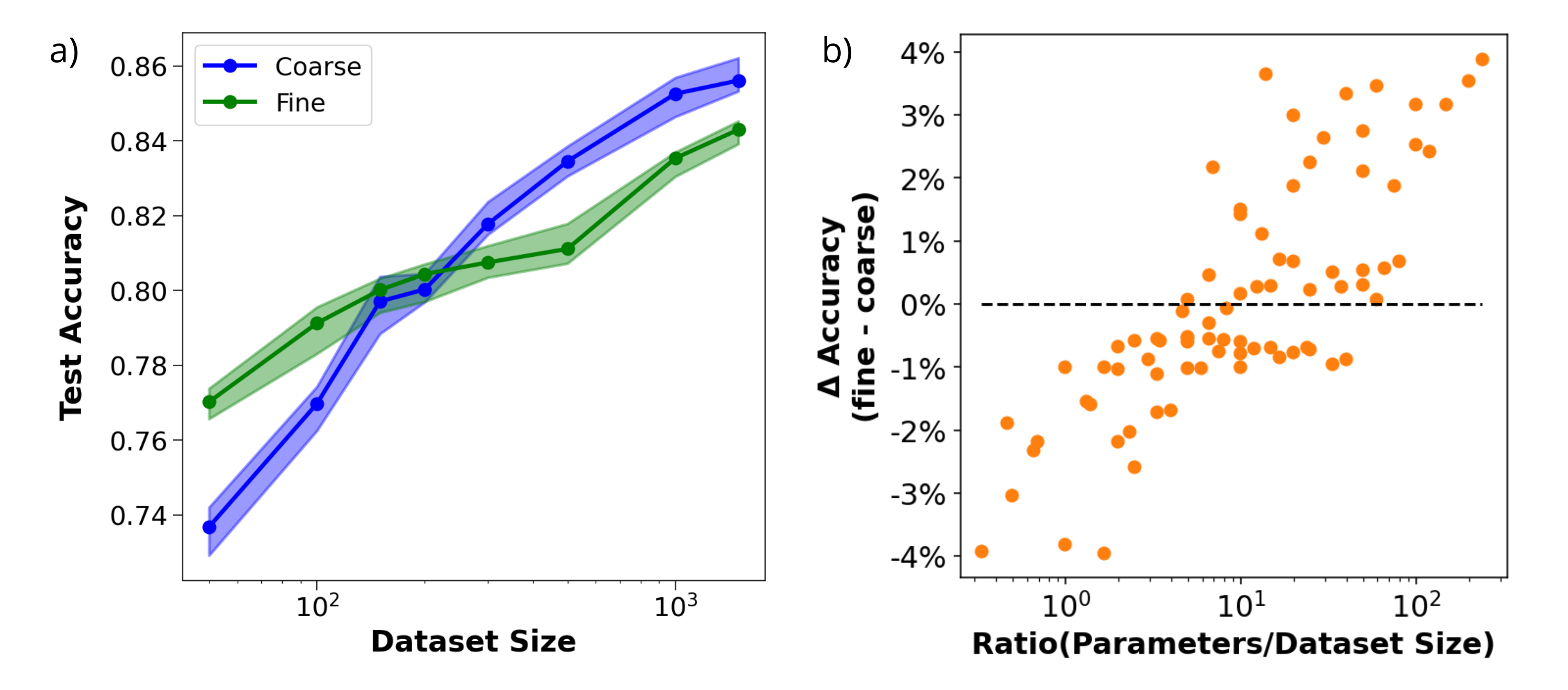}
    \caption{\textbf{Test accuracy with tanh activation function for the K-MNIST dataset.} a) Test accuracy on K-MNIST as a function of training set size for neural networks trained on fine-grained labels (green) and on coarse-grained labels (blue). Networks with 10 hidden neurons, trained with stochastic gradient descent (SGD) and early stopping. The batch size scales with dataset size (8 for the smallest, 32 for the largest). The learning rate was scheduled to decrease linearly from 0.01 to 0.001. Test set of 104 points, with error bars representing the first and third quartiles of accuracy obtained from 30 independent training runs with different initializations. (b) The difference in test accuracy against the ratio of model parameters to dataset size for the various datasets. Each point represents the fine-grained model's accuracy gain (or loss) for a specific combination of dataset size and model capacity.
    }
\end{figure}
\end{appendices}
 
\end{document}